# Finding the Loops that Matter

*By: Robert Eberlein and William Schoenberg*

## Abstract

The Loops that Matter method (Schoenberg et. al, 2019) for understanding model behavior provides metrics showing the contribution of the feedback loops in a model to behavior at each point in time. To provide these metrics, it is necessary find the set of loops on which to compute them. We show in this paper the necessity of including loops that are important at different points in the simulation. These important loops may not be independent of one another and cannot be determined from static analysis of the model structure. We then describe an algorithm that can be used to discover the most important loops in models that are too feedback rich for exhaustive loop discovery. We demonstrate the use of this algorithm in terms of its ability to find the most explanatory loops, and its computational performance for large models. By using this approach, the Loops that Matter method can be applied to models of any size or complexity.

## Overview

The Loops that Matter approach to understanding the structural sources of model behavior described in Schoenberg et. al., (2019) relies on the comparison of the identified feedback loops (using a loop score metric) in a model. In order to make that comparison, the set of feedback loops to use for comparison needs to be known. For small models, all the feedback loops can be found, scored, and compared. For large and feedback rich models, this is not practical just because of the overwhelming number of potential feedback loops as discussed in the background section.

This paper focuses on finding the set of feedback loops to compare when doing loop dominance analysis using the Loops that Matter method. The loopset selection problem that has been solved before as described in Kampmann (2012), and Oliva (2004) but those solutions do not turn out to be appropriate for the purpose of finding the most explanatory loops. The second section in this paper, on Independent and Important loops demonstrates why this is the case. This demonstration is done using a simple arms race model, which is also helpful in understanding the approach we ultimately adopt in solving the search problem.

In the section on composite feedback structure, we will use a simple model for which the metrics of the Loops that Matter method (Schoenberg, et. al, 2020) can be computed by inspection. This simple model makes clear the dilemma of analyzing a complex model for which the importance of different links and loops is changing during the course of a simulation. As such, even though it is a trivial model, it provides a foundation for developing algorithms to discover important loops in complex models.

Once we have outlined where we want to go, we will discuss the path to get there. The development of the approach described in this paper has been a very iterative process, which many attempts showing promise abandoned in that development. In order to help future researchers who want to improve on our solution, we include discussion of several of those abandoned paths.

Finally we present the approach that we settled on for production use. The algorithm chosen is, heuristic. It does not guarantee the discovery of the truly strongest loops, but we have reason to believe that it is likely to discover strong loops, and observations on the nature of strong loops in large and complex models give us confidence they will be similar to, if not, the strongest loops.

Ironically, almost all of the discussion in this paper will be based around very simple models for which complete enumeration of feedback loops (a problem long ago solved by Tarjan (1973) and others) is trivial. While we will discuss the application of the techniques to larger models such as the Urban Dynamics model (which has 43,722,744 feedback loops) the main concepts are most intuitively understood with small examples.

## Background

Understanding the connection between structure and behavior requires both the recognition of system physics as shown in stock and flow diagrams, and the identification of feedback loops that are responsible for generating the behavior of interest. For small models identifying feedback loops is straightforward as it can typically be done visually and certainly computationally with little difficulty. For larger models, however, the number of feedback loops grows very quickly (potentially proportional to the factorial of the number of stocks) and can't be enumerated with any practicality (Kampmann, 2012). Since the purpose of finding the loops is to identify those that are important in generating behavior, we want to find an approach that identifies the loops most important to the observed behavior, and skips over those of less interest.

This difficulty is well recognized. Kampman (2012) makes it clear and suggests a method for solving it using what is termed an independent loop set that contains a number of loops which is typically proportional to the number of stocks. Oliva (2004) refines this by defining a shortest independent loop set that is both unique and easily discoverable. These solutions are based upon a static analysis of the model equations, and do not take into consideration the behavior which is produced by the model. While the independent loop set approach solves the identification problem in large models, it will not necessarily find the loops most important to understanding behavior as pointed out in Güneralp (2006) and Huang et al (2012). This is discussed further in the next section.

An alternative approach to discovering loops is to use a method that finds the more important loops first, and stops when the set of important loops have been found. This is what we describe in this paper; the method is a heuristic that gives good, but demonstrably incomplete, enumeration of important loops.

We base our discovery approach on the Loops that Matter technique for quantifying the contribution of loops outlined in Schoenberg et. al. (2019). This method is based on a link score that determined the contribution of a changing input to a changing output in an equation. The score for a loop, is then the product of all the link scores. This means that as we attempt to discover loops, we can know the score of each loop we find as soon as we detect it. By prioritizing search based on link scores, our discovery approach completes in reasonable time even for large and highly interconnected models.

## Independent and Important Loops

We use the simple, three-party arms race model shown in Figure 1 to demonstrate what an independent loop set (Kampman, 2012) is. This model also highlights why using the independent loop set will not always find the important feedback loops following the example of Huang et al (2012). Finally, we use this simple model to show how the loop discovery process can be sped up using its local characteristics.

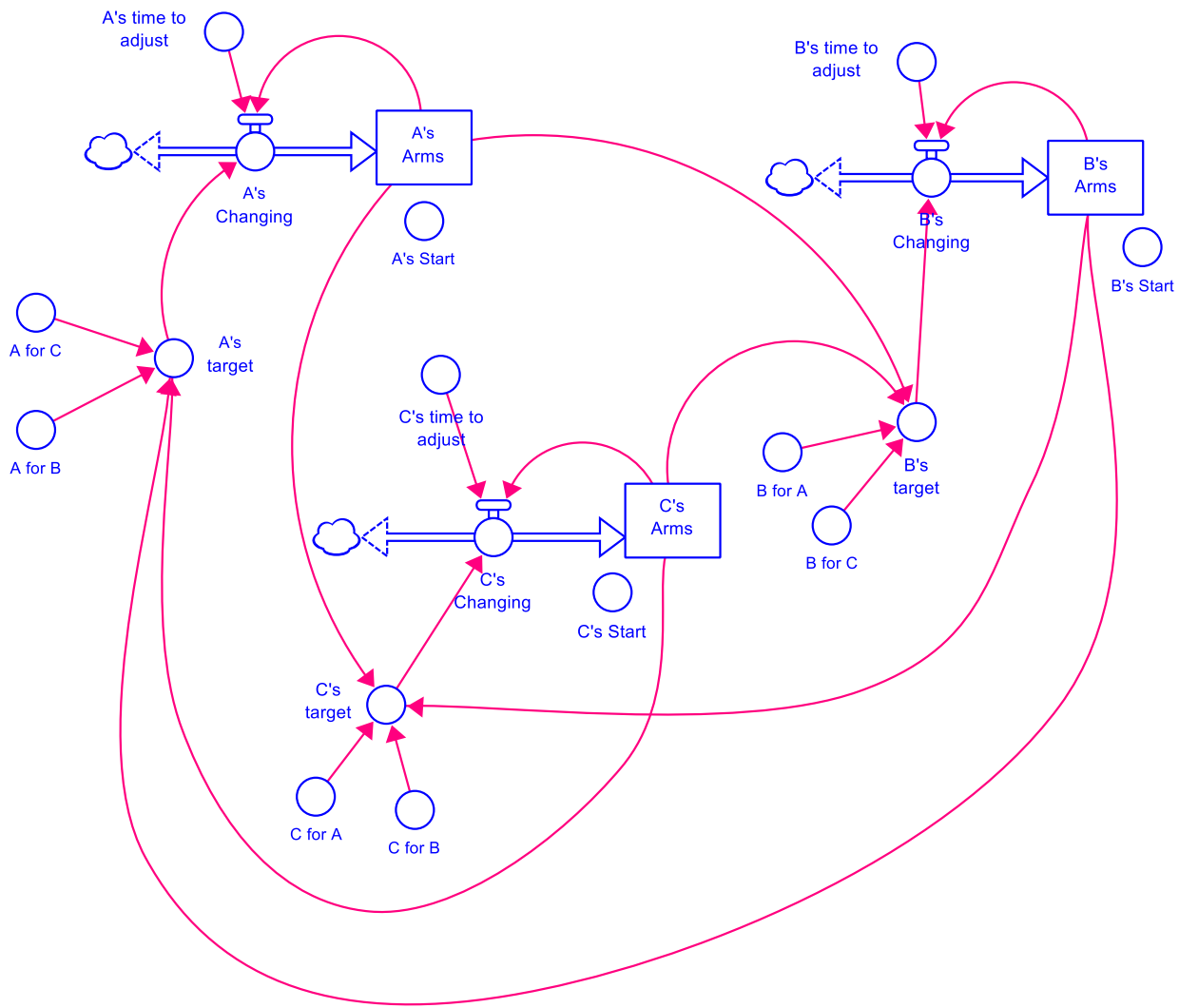

*Figure 1: A simple three-party arms race model.*

The model is set up so that A wants only parity with B and 90% of C, B wants parity with A and 110% of C, and C wants 110% of A and 90% of B. A starts at 50, B at 100 and C at 150. There are 3 balancing stock adjustment loops (the standard balancing loop in the arms race archetype), three pairwise reinforcing loops A, to B's target, to B, to A's target and so on (the standard reinforcing loops in the archetype) and then two reinforcing loops involving all three players (A to B's target to B to C's target to C to A's target and A to C to B (with intermediate)).

Following the terminology of Oliva (2004) and as identified by Huang et. al. (2012) in a similar model, the shortest independent loops would consist of the 3 stock adjustment loops, and the 3 pairwise reinforcing loops. The completeness of this set of loops is demonstrated by the fact that all of the connections involved in all of the feedback loops are used in this set of loops. The longer loops connect A to B (already used in the AB pair) and then B to C (already used in the BC pair) and then C to A (already used in the AC pair).

However, simulating this model, or even just thinking through the relative gains from the description above, it should be clear that the pairwise reinforcing loops all have gain equal to or smaller than one. Thus, focusing only on those loops, the behavior will necessarily be adjustment toward balance or toward zero. Because the shortest independent loops include all of the connectors, breaking any loop is guaranteed to break one of the shortest independent loops. This completeness, while guaranteeing we will see an effect from any change in the connection strength from one link to another, does not mean we will see the most important effect. The A to B to C loop is the one that determines the long term behavior of this model as shown in Figure 2, though it does not during the short term adjustments at the beginning.

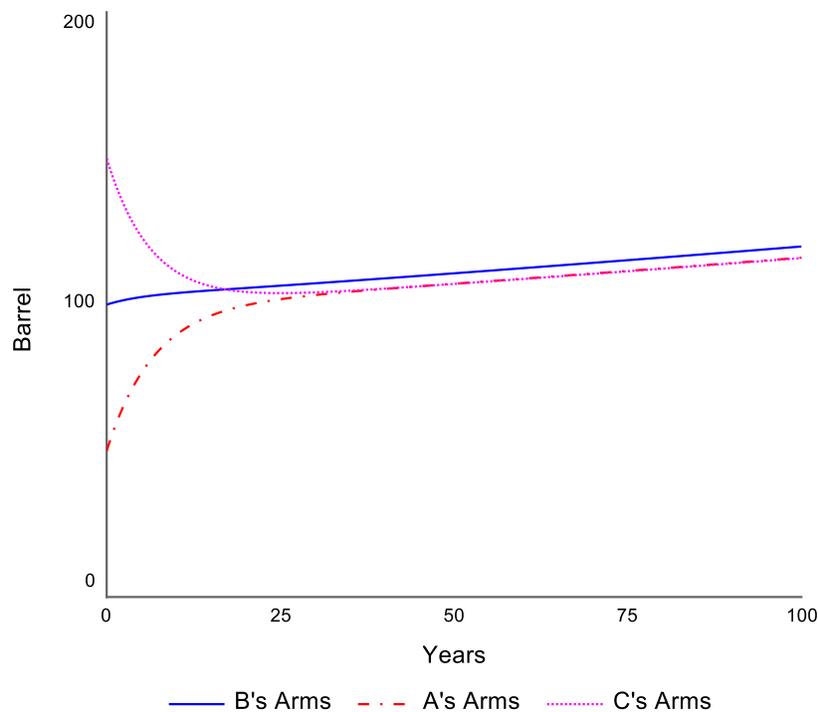

*Figure 2: Behavior of the 3 states in the arms race model.*

Which loop is important when can be seen quite easily using the loop score metric from the Loops that Matter method. Figure 3 shows the percent of behavior each loop is responsible for:

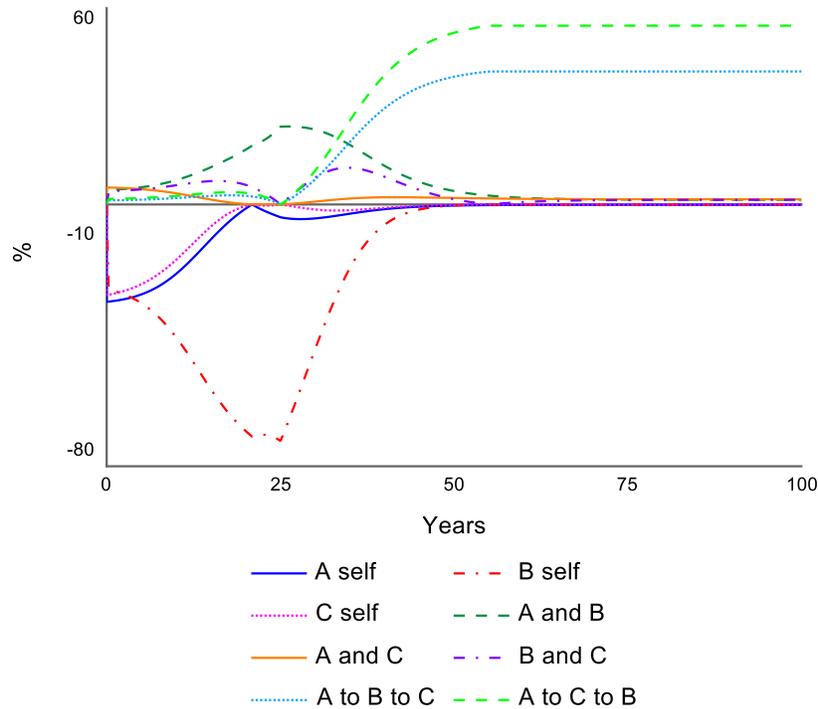

Figure 3: Relative loop scores for all loops in the arms race model

The paired interactions and self-corrections are important at the beginning. This is not surprising as we started the model quite far from a balanced trajectory. By about time 50 however, the model behavior is dominated by the two long loops.

For this model it is clear that we need to identify all of the feedback loops for analysis, as all of them (with the possible exception of the A and C interaction) are of consequence at some point in the simulation.

Once we have found the loops to analyze, the loop scores can be used to determine which loops are most explanatory. Until we find the loops, however, we can't apply the technique. As is clear from this example, we need to find things that will have different importance at different points in time, but we need to find everything that is ever important precisely to see which loops are driving behavior when.

## The Composite Feedback Structure

When we look at a model and try to find feedback loops we do so by following links and flows from one variable to another. Assuming the model diagram is accurate, and we are patient and assiduous, this will define the universe of potential loops. This approach, however, includes loops that may never be active during the course of a simulation. For example, consider a very simple, if contrived, two stock model shown below in Figure 4:

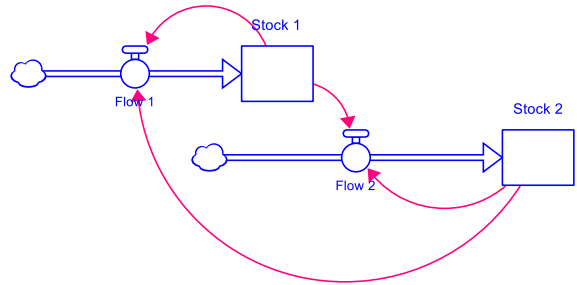

*Figure 4: Simple model showing decoupled feedback loops.*

Where

Flow_1 = IF Stock_2 > 50 THEN Stock_2/DT ELSE Stock_1/DT

And

Flow_2 = IF Stock_1 > 10 AND Stock_1 < 20 THEN Stock_1/DT ELSE Stock_2/DT

With both stocks starting at 1, until Stock_1 gets to 10 we just have the two minor loops running side by side. When Stock_1 is between 10 and 20 it drives Flow_2, but Flow_1 is determined only by Stock_1. Later, when Stock_2 finally reaches 50, it drives both flows. So what we have are two decoupled loops that alternatively drive the other flow. When that happens, there is only one feedback loop active as can be seen from in the Figure 5.

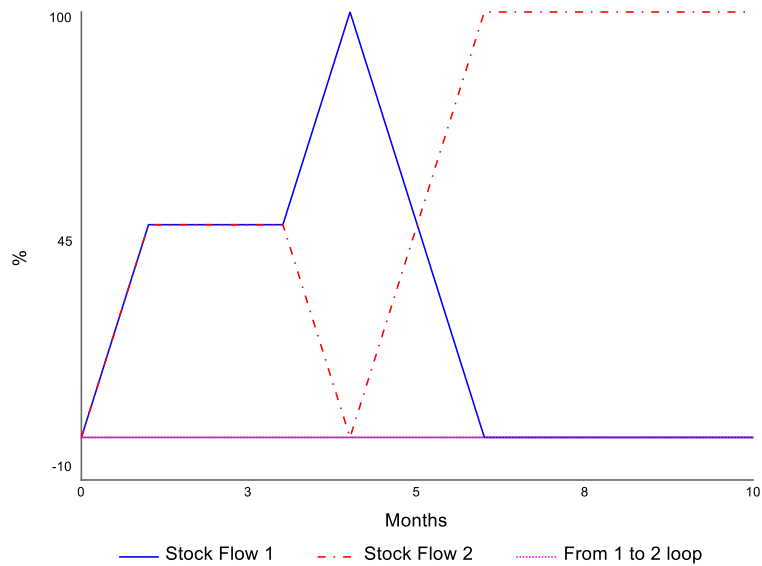

*Figure 5: Behavior of the simple model shown in Figure 4*

If, however, we were to measure the link scores for the components of the long loop they would look like this (Figure 6):

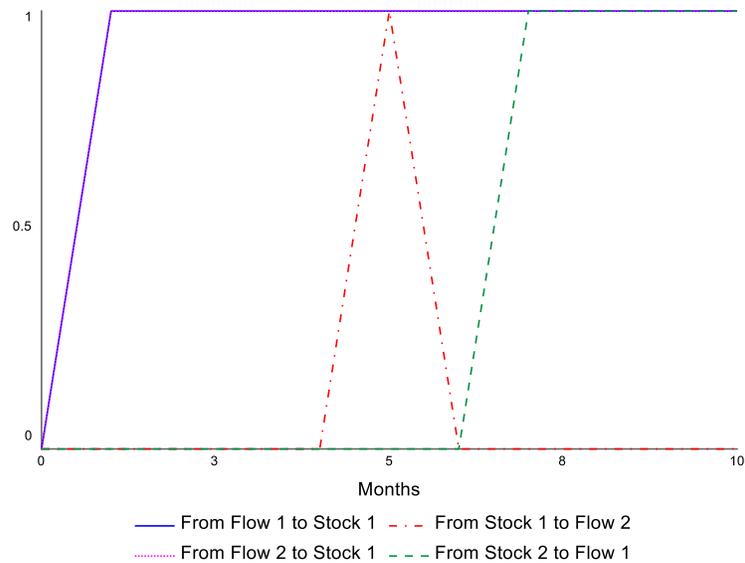

*Figure 6: Link scores of all links from the model shown in Figure 4*

The flow to stock connections are always 1, since there is only a single flow for each stock. All the link scores start at 0 because nothing has changed at the beginning of the simulation – a convention of the Loops that Matter scoring technique. The interesting things are the link from Stock_1 to Flow_2 is 1 at time 5, but otherwise 0 whereas the link from Stock_2 to Flow_1 is 0 until time 7 then 1 afterward (the dt in this model is 1).

In order to discover all of the feedback loops in a model, we need to include all links that are active at some point during the simulation. In this model, every link is active at some point in time, but not active at others. It is very common for links to become more and less active, though not necessarily completely inactive. As a consequence of this some loops are stronger at some times than others. To capture all the potential loops in this case, or the loops that could be strong at some point in time in the more general case, we need to pay attention to links that are strong at any time during the simulation. We do this by creating a composite link score that represents the contribution of a link over the course of the simulation. As long as the link is non-zero we will find all of the loops, and if we can choose the composite score well, it seems reasonable we will be able to identify the important loops.

One approach to creating the composite link scores to support loop discovery is to use the largest magnitude of the link score over all times. In this case, that would give us 1 for every link. Starting from either stock the connection from the stock's own flow, or the other stock's flow would be equally strong. Assuming we found all three feedback loops, they would all have a loop score of 1. Using this approach, the composite loop scores will always be as big, or bigger, than any actual loop score. This can be a problem for a big model, where the loop scores can become extremely large, and when using the largest link score magnitudes, the longer the loop the bigger the score. This approach is biased to finding longer loops, and also suffers from numeric problems because the size of the composite scores can easily exceed 1.0E300.

An alternative to approach to using the largest link score magnitude, is to take an average of the link score magnitude instead. In this case the flows would average to 1, while the link from Stock_1 to Flow_1 would average 0.5, from Stock_2 to Flow_2 0.9, from Stock_1 to Flow_2 0.1, and from Stock_2 to Flow_1 0.5. The strongest links out of either stock would be to its own flow. The minor loop for Stock_1 would get score 0.5, the minor loop for Stock_2 0.9 and the major loop 0.05. At first glance, this seems like a reasonably approximation since the minor loops are always 1 or 0 and the major loop is always 0. The problem is that with a small change to the equations, the major loop could also have the same 1 or 0 behavior. Thus, with averages, it is generally true that the longer the loop, the lower the score. This approach is biased toward finding shorter loops, but does have nice numeric properties as the scores computed do not tend to grow abnormally large.

We spent a great deal of time trying to make both of these approaches to building a composite network converge, but ultimately the detected loops that ranked as most important, simply were not the most important loops when looking at the scores over the full simulation. Ultimately, we adopted the obvious solution, which is to do loop detection at every (or almost every) point in time. This solution does require many more discovery passes, but the strongest path algorithm used for each pass converges much more quickly than the searches that were required on the composite network making each pass much faster.

The composite feedback structure is used to perform an initial identification attempt that will find loops exhaustively if there are not too many (less than 1000). For this the actual link scores selected do not matter, but we use the maximum of all link scores in the computation. Many models have less than 1,000 loops and for these we don't need to go through the identification process described in the next sections.

## The Strongest Path Technique

We set out to find an algorithm that was similar in spirit to the shortest path algorithm of Dijkstra (1959), derivatives of which form the bases for route selection in modern GPS systems. At its heart Dijkstra 's algorithm is simple. If the intent is to go from a to b, and while investigating possible routes we pass through c after 10 kilometers, then any other route that passes through c in more than 10 kilometers is not worth pursuing. This way the entire set of possible paths from a to b do not need to be explored.

The most direct analog to the shortest path search in the context of loop discovery would be to start from a and try to find paths to b, and instead of tracking the distance, which is additive, use the link score magnitude which is multiplicative. Then, when we get to c we check if we have been to c before and if we have, whether the score we got on that path was bigger. If it was, we don't explore further.

Unfortunately, since we are trying to maximize instead of minimize, this approach does not lead to an exact solution. Specifically, if the path we took to c won't actually get us back to a, but instead will form a loop not involving a, then the score may be bigger than what we would get on the strongest path from a to a. This is easily seen in Figure 7:

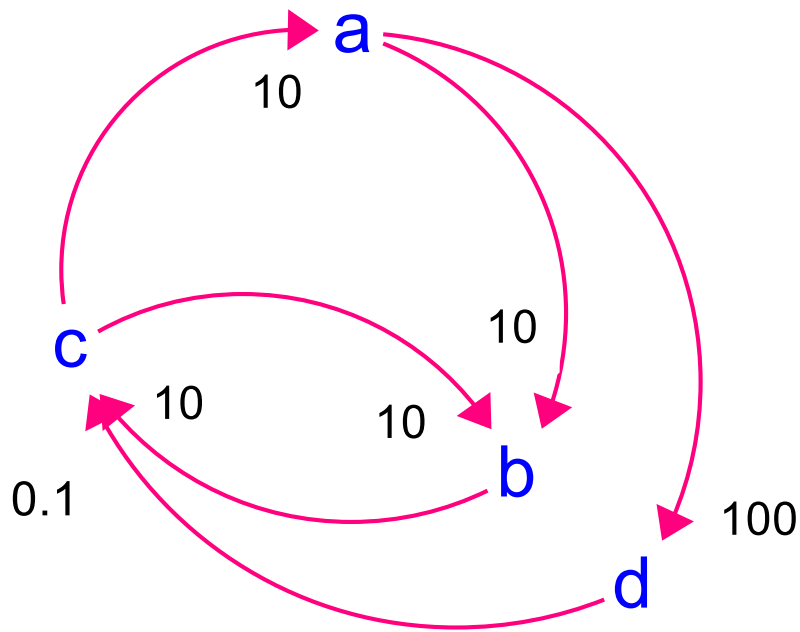

*Figure 7: Demonstration of a failure case in directly applying Dijkstra's algorithm to loop finding*

The numbers are the link scores. Starting from a and going to d gives us a score of 100, then to c gives us 10, then to b 100. Starting from a and going directly to b would give us a score of 10, which is less than 100 so we would not pursue that path. Thus, we would not find the loop a->b->c->a which has a score of 1000. Instead we would find a->d->c->a which has a score of 100.

In the above diagram, starting at b could give us the a->b->c->a loop, and starting at c might. It is however, possible (though messy) to create diagrams where starting anywhere would fail to find the strongest loop.

With those caveats in place, we can describe our loop discovery algorithm, this is repeated for each computational interval or a subset of those based on performance tuning.

1. Compute the link scores for every connection in the model. Some of these may be 0.
2. For every variable in the model sort the outbound links (places where the variable is used) by the link scores so that the first link has the biggest (absolute value) score.
3. For every stock in the model (all loops involve a stock) start the search:
    a. Go through each outbound link in order, multiply by the score of that link then test the variable the link points to with the score
        i. If the variable is the starting variable record the loop and the score
            1. Need to check loop for uniqueness. If we already have it, then ignore it.
        ii. If the variable is being visited already (a loop not involving the starting point) just return as the loop will be found starting from another stock.
        iii. If the variable has been visited, and has a higher score from that visit just return.

iv. If the variable has not been visited, or has a lower score, mark the new score and execute step a. above.

The algorithm has a similar computational burden to the shortest path algorithm (roughly proportional to the square of the number of variables). Sorting the edges by score helps to increase efficiency as it makes it more likely that the first visit to a variable will be the one with the highest score. Most of the computational burden ends up being in the determination of uniqueness and the construction of the loop information for later processing.

Pseudo-code for the algorithm that is slightly more formal than the above outline is included in Appendix 1.

## Completeness

It is possible to compare the loops found using the strongest path algorithm with an exhaustive search of loops for models with a relatively small (less than 100,000) set of unique loops. We can then compare the loops found using our approach to the full set of loops, in both cases sorting them by the average contribution of the loop to behavior over the course of the simulation.

For the Market Growth Model from Forrester (1968, as replicated in Morecroft (1983)) there are a total of 19 loops. When the loop discovery algorithm is run on this model it discovers all 19. A good result, so there is not much more to show.

For the service quality model of Oliva and Sterman (2001) there are a total of 104 loops in the main set of loops, though only 38 have a contribution of more than 0.01% to behavior based on the Loops that Matter contribution metrics. There are also 4 additional sets of loops which are smoothed quality measures that do not get used elsewhere in the model (a smooth is a single negative feedback loop). When loops are discovered using the strongest path algorithm a total of 76 loops are discovered with 28 having a 0.01% or greater contribution to behavior. Of the first 15 loops in the full set, the eighth is not present in the set generated by the strongest path algorithm. Table 1 compares the 8th loop in the full set to the 4th in both sets:

Table 1: Comparison of the 4th and 8th loops from the strongest path algorithm found in the Service Quality model

| 4th Loop | 8th Loop |
| --- | --- |
| experience rate | experience rate |
| Experienced Personnel | Experienced Personnel |
| total labor | total labor |
| on office service capacity | on office service capacity |
| service capacity | service capacity |
| work pressure | work pressure |
| work intensity | work intensity |
| potential order fulfillment | potential order fulfillment |
| order fulfillment | order fulfillment |
| Service Backlog | Service Backlog |
| desired service capacity | desired service capacity |
| Change Desired Labor | Change Desired Labor |
| Desired Labor | Desired Labor |
| labor correction | labor correction |
| desired hiring | desired hiring |
| desired vacancies |  |
| vacancies correction |  |
| indicated labor order rate | indicated labor order rate |
| labor order rate | labor order rate |
| Vacancies | Vacancies |
| hiring rate | hiring rate |
| Rookies | Rookies |

The two loops are identical through desired hiring, then the same again starting from indicated labor order rate. Indeed this type of miss is quite typical. Large models have many loops that are very similar, but not identical. The strongest path algorithm will often miss one of the shorter or longer loops with a similar set of elements.

Another interesting model is the economic cycles model from Mass (1975). This model has a total of 494 feedback loops (again there are some other set of loops in addition to the main set). When run through the strongest path algorithm there are 261 loops discovered. Out of the first 40 loops, only the 22$^{nd}$ and 40$^{th}$ are missing. Again the missing loops have much in common with the loops around them, though they are not quite as simply related as the service quality example.

## Performance

Much of the work in the development of these algorithms used the Urban Dynamics model from Forrester (1969). When fully enumerated, this model has 43,722,744 loops. This is too big a number to compare the algorithm with the fully enumerated set. The determination of the loops is fairly fast, 10 to 20 seconds on an 8$^{th}$ generation intel core I7 processor. The strongest path algorithm discovers 20,172, though this is truncated to less than 200 for display, by only choosing those loops which describe at least 0.1% of the total behavior of the model when compared against all loops found at the termination of the search.

Even though we can't make definitive comparison against the full set of loops in the Urban Dynamics model, we can compare the results of different loop discovery approaches. We have experimented on this model with a number of different algorithms for loop discovery and saw the same pattern we highlighted for the service quality example. When an algorithm was tuned to run faster (and find fewer loops), the missing loops were quite similar, but somewhat longer or shorter, to those common to both tunings of the algorithm. Of course, with such a large set of loops, we can never know if we found the loop with the biggest score, but we have some confidence that we will find a loop that looks quite similar.

Mostly out of curiosity we also did a quick analysis of the World 3-03 model by Meadows et al (2004). This model has 330,574 loops when fully enumerated. Not nearly as many as the Urban Dynamics model, but still too many to look at. The algorithm finds a total of 2,709 loops and this is truncated to 112 loops using the 0.1% contribution cutoff, with a number of both reinforcing and balancing loops being identified as important at different phases in the simulation. This computation takes about 4 seconds.

## Paths not Taken

Jay Forrester often said that if we really want people to learn from us we should write about our failures. The development of the strongest path technique was the result of many attempts using different approaches that ultimately proved unsatisfactory.

The one thing that we got right from the beginning was the ordering of search nodes by their strongest connection. This seemed such an intuitive thing, that we never tested it, except by accident, and when we did the results were as expected. Ordering in this manner often speeds up search times by a factor of 3.

The initial attempts at getting algorithms to converge in reasonable time were focused on the remaining potential given the path that had been traversed, and were computed on the composite network. The first attempt at this was to trace the strongest links out of a node until we hit a terminal node or detected a loop and then use the biggest score found along that path as the potential of a node (variable) to add to a loop. Then, when the current score potential for the loop being discovered fell below a threshold (set based on loops already discovered) the search would be abandoned. Ultimately, the problem with this approach was that picking only the strongest outbound link seems at best modestly correlated with the real potential. The other problem is that the forward path might already include variables that have been visited.

To guard against counting the value of links already visited, we developed the concept of the total potential score remaining. This was computed as the product of the score of the strongest link out of every variable. As the variables are traversed looking for loops, the potential is decreased by the node being investigated, while the current score is increased by the actual path out of the variable. The product of the current score and potential is thus monotonically decreasing. Picking an appropriate threshold at which to terminate the search (again based on loops already found) then provides a way to stop pursuing paths unlikely to uncover high scoring loops. This approach actually worked very well for composite scores that were based on the average of the scores over the simulation. The trouble with this approach was that the most powerful loops identified this way did not correlate at all with the most powerful loops as determined by looking at the total contribution of the loops to the dynamics. When we used the biggest values for the composite scores, the numbers were simply too large to provide reasonable cutoff thresholds.

A completely different approach we tried was to trim the feedback structure by getting rid of links. The expectation was that this would allow us to exhaustively search loops on a less connected model. The first attempt to do this removed links after they had been included in a requisite number of loops, based on the assumption that the strongest loops were found first.  The second attempt to do this removed all the weak links.  The difficulty with this approach was that the links being pulled out might be necessary to complete a loop with a high score, even though the link itself was not scored highly. This is not necessarily a fatal flaw in the approach, as we have seen that loops typically have siblings with a few more or fewer links. We were not, however, able to convince ourselves that this was all that was missing, nor were we sure of performance in models of different size.

We also explored the compaction of the model using only stock to stock connections. This has some theoretical appeal. Since all loops involve stocks, it seems like this would give us a smaller network to search and therefore improve performance. We also felt that the stock to stock connections could be selected for maximum strength and this would give more accurate results. As it turns out, computational speed improvements did not really occur. Ultimately, it is really the number of paths that exist and we don't change that much by eliminating non stocks, but instead just make more connections between the variables that do remain. Worse, when we do eliminate parallel paths, we have dropped potential feedback loops in the model. These might be more easily interpretable than the ones selected, and including both of them might be informative.

## Conclusions

We have demonstrated the importance of drawing from all loops when determining the loops that matter, and presented a technique for finding the most important set of loops using the strongest path algorithm that is practical and gives good results in large models. Though it is a heuristic, we feel that it gives a good balance of outcome and computational burden and has shown its utility for the models we have experimented with.

Ultimately, the value of the technique depends on its utility for providing understanding to model builders and consumers. As it is being embedded in a comprehensive framework for model analysis it is likely that users will find new needs relative to the discovery and display of loops.

## Appendix I Algorithm Pseudo-Code

```
STACK – a vector of variables
TARGET – the stock currently under investigation
Function Check_outbound_uses(variable,score)
    If variable.visiting is true
        If  variable = TARGET
            Add_loop_if_unique(STACK, variable)
        End if
        Return
    End if
    If score is less than variable.best_score
        Return
    End if
```

```
            Set variable.best_score = score
            Set variable.visiting = true
            Add variable to STACK
            For each link from variable
                Call Check_outbound_uses(link.variable, score*link.score)
            End for each
            Set variable.visiting = false
            Remove variable from STACK
End function
For each time in the run
      For each variable in the model
            For each link from variable
                  Set link.score from available data on link score
            (outside scope)
            End For each
            Set variable.best_score = 0
      End For eash
      For each stock in the model
            Set TARGET = stock
            Check_outbound_uses(stock, 1.0)
      End for
End for
```